\documentclass[sigconf,nonacm]{acmart}

\usepackage{balance}

\setcopyright{rightsretained}

\acmSubmissionID{1068}

\usepackage{amsmath}
\usepackage{amssymb}
\usepackage{amsbsy}
\usepackage{mathtools}

\usepackage{xspace}
\usepackage{subfigure}
\usepackage{booktabs}
\usepackage{graphicx}  
\usepackage{tabularx}
\usepackage{multirow}
\usepackage{listings}
\usepackage{pgfplots}
\pgfplotsset{compat=1.3}
\usepackage{array}
\newcolumntype{C}[1]{>{\centering\let\newline\\\arraybackslash\hspace{0pt}}m{#1}}
\newcolumntype{L}[1]{>{\raggedright\let\newline\\\arraybackslash\hspace{0pt}}m{#1}}
\newcolumntype{R}[1]{>{\raggedleft\let\newline\\\arraybackslash\hspace{0pt}}m{#1}}
\usepackage{amsthm}
\theoremstyle{definition}

\usepackage{soul}
\usepackage{paralist}
\usepackage{algorithm}
\usepackage{algpseudocode}

\renewcommand{\algorithmiccomment}[1]{\bgroup$\triangleleft$~{\color{blue}\textit{#1}}\egroup}

\newcommand{\ie}{i.e.\@\xspace}
\newcommand{\eg}{e.g.\@\xspace}
\newcommand{\cf}{cf.\@\xspace}
\newcommand{\ca}{ca.\@\xspace}

\newcommand{\wrt}{w.r.t.\@\xspace}

\newcommand{\kg}{\ensuremath{\mathcal{G}}\xspace}

\newcommand{\Real}{\ensuremath{\mathbb{R}}\xspace}
\newcommand{\Complex}{\ensuremath{\mathbb{C}}\xspace}
\newcommand{\set}[1]{\ensuremath{\mathcal{#1}}\xspace}
\newcommand{\vv}[1]{\ensuremath{\pmb{#1}}\xspace}

\newcommand{\ent}[1]{\textit{#1}}
\newcommand{\rel}[1]{\textit{#1}}

\newcommand{\triple}[3]{\ensuremath{( #1, #2, #3 )}\xspace}

\newcommand{\tprod}[3]{\ensuremath{\langle #1, #2, #3 \rangle}\xspace}
\newcommand{\params}{\ensuremath{\Theta}\xspace}
\newcommand{\score}[2]{\ensuremath{\phi_{r}(#1, #2)}\xspace}
\newcommand{\fscore}{\ensuremath{\phi}\xspace}
\DeclarePairedDelimiterX{\norm}[1]{\lVert}{\rVert}{#1}
\newcommand{\topic}[1]{\vspace*{1em}\noindent{\bf #1.\@\xspace}}
\newtheorem*{example}{Example}
\newcommand{\bmin}{\varphi^{\downarrow}}
\newcommand{\bmax}{\varphi^{\uparrow}}
\newcommand{\best}{\cellcolor{YellowGreen!60}}
\newcommand{\bb}[1]{\bf #1}
\newcommand{\ourtitle}{Embedding Cardinality Constraints in Neural Link Predictors}

\usepackage[capitalise]{cleveref}
\newtheorem{definition}{Definition}

\begin{document}
%
\title{\ourtitle}

\author{Emir Mu\~{n}oz}
\orcid{0000-0002-0089-8135}
\affiliation{%
  \institution{Data Science Institute, \\National University of Ireland Galway}
  \streetaddress{Galway Business Park, Dangan}
  \city{Galway}
  \country{Ireland}
  \postcode{H91 AEX4}
}
\email{emir@emunoz.org}

\author{Pasquale Minervini}
\orcid{0000-0002-8442-602X}
\affiliation{%
  \institution{University College London}
  \city{London}
  \country{United Kingdom}
}
\email{p.minervini@cs.ucl.ac.uk}

\author{Matthias Nickles}
\orcid{0000-0002-9194-6197}
\affiliation{%
  \institution{Data Science Institute, \\National University of Ireland Galway}
  \streetaddress{Galway Business Park, Dangan}
  \city{Galway}
  \country{Ireland}
  \postcode{H91 AEX4}
}
\email{matthias.nickles@nuigalway.ie}

\renewcommand{\shortauthors}{E. Mu\~{n}oz et al.}

\begin{abstract}
Neural link predictors learn distributed representations of entities and relations in a knowledge graph.
They are remarkably powerful in the link prediction and knowledge base completion tasks, mainly due to the learned representations that capture important statistical dependencies in the data.
Recent works in the area have focused on either designing new scoring functions or incorporating extra information into the learning process to improve the representations.
Yet the representations are mostly learned from the observed links between entities, ignoring commonsense or schema knowledge associated to the relations in the graph.
A fundamental aspect of the topology of relational data is the cardinality information, which bounds the number of predictions given for a relation between a minimum and maximum frequency.
In this paper, we propose a new regularisation approach to incorporate \emph{relation cardinality constraints} to any existing neural link predictor without affecting their efficiency or scalability.
Our regularisation term aims to impose boundaries on the number of predictions with high probability, thus, structuring the embeddings space to respect commonsense cardinality assumptions resulting in better representations.
Experimental results on Freebase, WordNet and YAGO show that, given suitable prior knowledge, the proposed method positively impacts the predictive accuracy of downstream link prediction tasks.
\end{abstract}


%
%


\copyrightyear{2019} 
\acmYear{2019} 
\setcopyright{acmcopyright}
\acmConference[SAC '19]{The 34th ACM/SIGAPP Symposium on Applied Computing}{April 8--12, 2019}{Limassol, Cyprus}
\acmBooktitle{The 34th ACM/SIGAPP Symposium on Applied Computing (SAC '19), April 8--12, 2019, Limassol, Cyprus}
\acmPrice{15.00}
\acmDOI{10.1145/3297280.3297502}
\acmISBN{978-1-4503-5933-7/19/04}

\maketitle

\section{Introduction}\label{sec:introduction}

%
Cognitive development of children indicates that we learn the cardinality-related question \textit{``How many?''} at ca. 3.5 years of age~\cite{ISI:A1990DW01300003}.
This ability helps us to recognise physical and abstract things by counting.
For example, a hand has commonly five fingers, a car has four wheels, or a meeting has at least two participants.
This kind of common sense knowledge is not obvious for machines to acquire, even in contexts where it can be useful, such as Question Answering, Web Search, and Information Extraction~\cite{DBLP:journals/sigmod/TandonVM17}.
One fundamental application area for cardinality information relates to the completion of Knowledge Graphs (KGs), graph-structured knowledge bases where factual knowledge is represented in the form of relationships between entities.
For instance, consider Freebase~\cite{DBLP:conf/aaai/BollackerCT07}, the core of the Google Knowledge Graph project, where 71\% of the people described in it have no known place of birth as reported by~\citet{DBLP:conf/kdd/0001GHHLMSSZ14}.
By leveraging cardinality information about the \rel{bornIn} relationship (\ie, each person must have a place of birth), we can quantitatively assess the degree of incompleteness in Freebase and focus the resources on predicting a single place of birth for each person.
Yet \emph{link prediction models} aimed at identifying missing facts in KGs do not consider such commonsense or schema knowledge, yielding potentially inconsistent and inaccurate predictions.
In this work, we focus on a certain class of link prediction models, namely \emph{Neural Link Predictors}~\cite{DBLP:journals/pieee/Nickel0TG16}.
Such models learn low-dimensional distributed representations---also referred to as \emph{embeddings}---of all entities and relations in a knowledge graph.
Neural link predictors are currently the state of the art approach to tasks such as link prediction~\cite{DBLP:conf/nips/BordesUGWY13,DBLP:journals/corr/YangYHGD14a,DBLP:conf/icml/TrouillonWRGB16,DBLP:conf/acl/WangWGD18}, entity disambiguation and entity resolution~\cite{DBLP:journals/ml/BordesGWB14}, taxonomy extraction~\cite{DBLP:conf/www/NickelTK12,DBLP:conf/nips/NickelK17}, and probabilistic question answering~\cite{DBLP:conf/semweb/KrompassNT14}.
Recently, research focused mainly on designing new scoring functions, and incorporating additional background knowledge during the learning process.
We refer readers to~\cite{DBLP:journals/pieee/Nickel0TG16,DBLP:journals/tkde/WangMWG17} for a recent overview on this topic.
In this paper, we address the problem of incorporating prior knowledge in the form of relation cardinality information into state-of-the-art neural link predictors.
For instance, we want to encode prior knowledge in the form of cardinality statements such as \textit{``a person should have at most two parents''} or \textit{``a patient should be taking between 1 and 5 drugs at a time''} in neural link prediction models.
Such prior knowledge can be provided by domain experts, or automatically extracted from data~\cite{DBLP:conf/wsdm/GalarragaRAS17,DBLP:conf/dexa/MunozN17}.
It is expected that such cardinality constraints will be satisfied by both the facts in the knowledge graph and algorithms analysing the graph, such as link predictors.
We believe that these constraints can impose commonsense knowledge upon the structure of the embedding space, thus helping us to learn better representations.
\begin{table}[t]
\centering
\resizebox{1.0\columnwidth}{!}{%
\begin{tabular}{lc}
\toprule
\multicolumn{1}{c}{\bf Triples}                        & {\bf Probability} \\
\midrule
\triple{edgar}{hasParent}{edgar}                       & 0.989  \\
\triple{edgar}{hasParent}{eliza\_poe}                  & 0.979  \\
\triple{edgar}{hasParent}{virginia\_eliza\_clemm\_poe} & 0.974  \\
\triple{edgar}{hasParent}{julia\_ward\_howe}           & 0.890  \\
\triple{edgar}{hasParent}{benjamin\_franklin}          & 0.889  \\
\bottomrule
\end{tabular}
}
\caption{Top-5 predictions (among 24 results with probability $>0.8$) for the \rel{hasParent} relation with \ent{Edgar Allan Poe} given by DistMult~\cite{DBLP:journals/corr/YangYHGD14a} on the FB13 dataset~\cite{DBLP:conf/aaai/BordesWCB11}.}
\label{tab:toy-example}
\end{table}
Cardinality constraints are one of the most important constraints in conceptual modelling~\cite[Chapter~4]{DBLP:books/daglib/0019156} as they explicit the topology of data.
However, existing neural link prediction models are not designed to incorporate them for learning better representations and more accurate models.
\begin{example}
One may expect that when predicting the parents (represented by relation \rel{hasParent}) for the entity \ent{Edgar Allan Poe}, a model will predict at most two parents, preferably \ent{Eliza Poe} and \ent{David Poe Jr.}
To illustrate this, let us analyse the actual predictions of a state-of-the-art neural link prediction model, DistMult~\cite{DBLP:journals/corr/YangYHGD14a}, using the Freebase FB13 dataset~\cite{DBLP:conf/aaai/BordesWCB11}, containing entities of the Freebase type \textit{deceased people} and their relations.
\Cref{tab:toy-example} shows the top-5 predicted parents for \ent{Edgar Allan Poe}.
As we can see, all predictions have a high probability (with 24 entities scored higher than $0.8$), albeit some predictions are incorrect.
\end{example}
Nevertheless, the evaluation results of our example model are positive due to the evaluation protocol of link prediction models based on a ranking metric, where correct predictions (\eg, \ent{eliza\_poe}) are expected to be ranked higher than incorrect ones (\eg, \ent{benjamin\_franklin}).
To address this problem, in this paper we propose an efficient approach for embedding the notion of cardinality in neural link prediction models, without affecting their efficiency and scalability.
The proposed approach is based on a novel regularisation term, that constraints the number of predictions for a given relation.
Briefly, our idea is to penalise the model when its predictions violate one cardinality constraints, expressed as lower or upper bound on the cardinality of a given relation type.
By doing so, the notion of cardinality of a relation will be captured during training, yielding to more accurate link prediction models, that comply with available prior knowledge~\cite{DBLP:conf/ijcai/WangWG15}, and learn better representations for entities and relations in the knowledge base.
\topic{Organisation}
The remainder of this paper is organised as follows.
First we present the definitions of knowledge graphs and neural link prediction models in~\cref{sec:background}.
Next we present the concept of relation cardinality constraint for knowledge graphs in~\cref{sec:cardinality}.
In~\cref{sec:regularizer}, we introduce a cardinality regularisation term which allows neural link predictors to leverage available cardinality constraints.
We evaluate the application of our regularisation term over different datasets and models in~\cref{sec:evaluation}.
\Cref{sec:rel-work} briefly discusses the existing works in link prediction over knowledge graphs.
Finally, \cref{sec:conclusions} concludes this paper. 
%

\section{Background}\label{sec:background}
%
We start by introducing the fundamentals of knowledge graphs and neural link predictors. 
%

\begin{definition}[Knowledge Graphs]
A \emph{knowledge graph} is a graph representation of a knowledge base.
Let \set{E} be the set of all entities, and \set{R} the set of all relation types (predicates).
We denote by \kg a knowledge graph comprising a set of \triple{h}{r}{t} facts or triples, where $h,t\in \set{E}$ and $r\in \set{R}$.
We refer to $h,t$ as \textit{subject} and \textit{object} entities and to $r$ as \emph{relation} of a triple.
Let $N_e = |\set{E}|$ and $N_r = |\set{R}|$ be the number of entities and relations, respectively.
\end{definition}

%
The goal of \emph{link prediction} models is to learn a scoring function \fscore that given a triple \triple{h}{r}{t} returns its corresponding \emph{score}, $\fscore\triple{h}{r}{t}\mapsto \Real$.
Such a score can then be used for ranking missing triples according to the likelihood that the corresponding facts hold true.

\begin{definition}[Neural Link Predictors]
\emph{Neural link prediction models}~\cite{DBLP:journals/pieee/Nickel0TG16,DBLP:journals/tkde/WangMWG17} can be interpreted as neural networks consisting of an \emph{encoding layer} and a \emph{scoring layer}.
Given a triple \triple{h}{r}{t}, the encoding layer maps entities $\ent{h},\ent{t}\in \set{E}$ to their $k$-dimensional distributed representations $\vv{e}_{h}$ and $\vv{e}_{t}$.
Then, the scoring layer computes the likelihood of the triple based on a relation-dependent function $\phi_{r}$.
Henceforth, the scoring function \fscore is defined as $\fscore\triple{h}{r}{t} = \score{\vv{e}_{h}, \vv{e}_{t}}$, where $\phi_{r}: \Real^{k} \times \Real^{k} \mapsto \Real$, $\vv{e}_{h}, \vv{e}_{t} \in \Real^{k}$, and $r\in \set{R}$.
\end{definition}

A neural link predictor with parameters $\params$ defines a conditional probability distribution over the truth value of a triple $\triple{h}{r}{t}$~\cite{DBLP:journals/pieee/Nickel0TG16}:
\begin{equation} \label{eq:model-prob}
p(y_{hrt} = 1 \mid \params) = \sigma(\score{\vv{e}_{h}}{\vv{e}_{t}}),
\end{equation}
\noindent where $y_{hrt}\in\{ 0, 1 \}$ is the truth label of the triple, $\params=\{\vv{e}_i\}^{N_e}_{i=1}\cup \{\vv{r}\}^{N_r}_{j=1}$ denotes the set of all entity and relation embeddings (the parameters \params), $\sigma(x) = 1/(1 + \text{exp}(-x))$ is the standard logistic function, and $\phi_{r}$ denotes the model's scoring function (\cf~\cref{tab:scoring-functions}).
Most models consider the $k$-dimensional embeddings as real-valued $\vv{e}_{h}, \vv{e}_{t}, \vv{r}_{r}\in \Real^{k}$; however, there are exceptions like ComplEx~\cite{DBLP:conf/icml/TrouillonWRGB16}, where $\vv{e}_{h}, \vv{e}_{t}, \vv{r}_{r}\in \Complex^{k}$.
A neural link prediction model is trained by minimising a loss function defined over a target knowledge graph $\kg$, usually using stochastic gradient descent.
Since knowledge graphs only contain positive examples (\ie facts), a way to provide negative learning examples---motivated by the Local Closed World Assumption (LCWA)~\cite{DBLP:conf/kdd/0001GHHLMSSZ14}---is to generate negative examples by \emph{corrupting} the triples in the graph~\cite{DBLP:conf/uai/RendleFGS09,DBLP:conf/nips/BordesUGWY13,DBLP:journals/pieee/Nickel0TG16}.
Given a (positive) triple $\triple{h}{r}{t} \in \kg$, corrupted triples (negative examples) can be generated by replacing either the subject or object with a random entity sampled uniformly from \set{E}~\cite{DBLP:conf/aaai/BordesWCB11}.
Formally, given a positive example $\triple{h}{r}{t}$, negative examples are sampled from the set of possible corruptions of $\triple{h}{r}{t}$, namely $\set{C}\triple{h}{r}{t} \triangleq \{\triple{h^\prime}{r}{t}\mid h^\prime \in \set{E} \} \cup \{\triple{h}{r}{t^\prime} \mid t^\prime\in \set{E} \}$.
Let $\set{D}^{+}$ be the set of positive examples, and $\set{D}^{-}$ the set of negatives generated accordingly with function \set{C}.
The training consists of learning the parameters $\params$ that best explain $\set{D}^{+}$ and $\set{D}^{-}$ according to~\cref{eq:model-prob}.
For that, models such as TransE~\cite{DBLP:conf/nips/BordesUGWY13}, DistMult~\cite{DBLP:journals/corr/YangYHGD14a} and HolE~\cite{DBLP:conf/aaai/NickelRP16} minimise a pairwise margin loss:
\begin{equation} \label{eq:hinge-loss}
    \mathcal{L}(\Theta) = \smashoperator[l]{\sum_{\tau^{+} \in \set{D}^{+}}} \smashoperator[r]{\sum_{\tau^{-} \in \set{D}^{-}}} \left[\gamma + \sigma(\fscore(\tau^{-})) - \sigma(\fscore(\tau^{+})) \right]_{+},
\end{equation}
\noindent where $\tau^{+} = \triple{h}{r}{t}$ is a positive example, $\tau^{-} = \triple{h^\prime}{r}{t^\prime}$ is a negative one, $[x]_{+} = \max(0, x)$, and $\gamma$ is the margin hyperparameter.
The entity embeddings are also constrained to unit norm, \ie $\forall i\in \set{E}: \norm{\vv{e}_{i}}_2 = 1$. 
Whereas other models like ComplEx~\cite{DBLP:conf/icml/TrouillonWRGB16} minimise the logistic loss:
\begin{equation*}\label{eq:logistic-loss}
 \mathcal{L}(\Theta) = \smashoperator[r]{\sum_{\tau\in \set{D}^{+}\cup \set{D}^{-}}} \text{log}(1 + \text{exp}(- y_{\tau} \fscore(\tau)))
\end{equation*}
\noindent where $\tau = \triple{h}{r}{t}$ is an example (triple), and $y_{\tau}\in \{-1, 1\}$ is the label (negative or positive) associated with the example.
\begin{table}[t]
\begin{center}
\resizebox{1.0\columnwidth}{!}{%
\begin{tabular}{lcc}
\toprule

\multicolumn{1}{c}{\bf Model} & {\bf Scoring Function} & {\bf Parameters} \\

\midrule

\multirow{2}{*}{ER-MLP} & \multirow{2}{*}{$\vv{w}^{T} \tanh \left( \vv{W}^{T} [\vv{e}_{h} ; \vv{e}_{t} ; \vv{r}_{r}] \right)$} & $\vv{r}_{r} \in \Real^{k}, \vv{w} \in \Real^{k'}$ \\

& & $\vv{W} \in \Real^{3k \times k'}$ \\

DistMult & $\tprod{\vv{e}_{h}} {\vv{r}_{r}} {\vv{e}_{t}}$ & $\vv{r}_{r} \in \Real^{k}$ \\

ComplEx & $\mathrm{Re}(\tprod{\vv{e}_{h}} {\vv{r}_{r}} {\overline{\vv{e}}_{t}})$ & $\vv{r}_{r} \in \Complex^{k}$ \\

\bottomrule
\end{tabular}
}%
\end{center}
\caption{Scoring functions \score{\vv{e}_{h}}{\vv{e}_{t}} of three state-of-the-art knowledge graph embedding models.}
\label{tab:scoring-functions}
\end{table}
%

\section{Relation Cardinalities}\label{sec:cardinality}
%
A relation type can have associated cardinality bounds, which restrict the number of object values that a subject can have.

\begin{definition}[Relation Cardinality Bound]
Let $\varphi_{r} = (\bmin_{r}, \bmax_{r})$ be a \emph{cardinality bound} for the relation $r\in \set{R}$, where $\bmin_{r}\in \mathbb{N}$ denotes the \emph{lower bound} and $\bmax_{r} \in \mathbb{N} \cup \{\infty\}$ denotes the \emph{upper bound} of the cardinality, s.t. $0\le \bmin_{r}\le \bmax_{r}$~\cite{DBLP:conf/dexa/MunozN17}.
A knowledge graph \kg \emph{satisfies} a cardinality bound $\varphi_{r}$ with $r\in \set{R}$ iff
\begin{equation*}\label{eq:satisfaction}
    \forall h\in \set{E}, (\bmin_{r}\leq count(r, h) \leq \bmax_{r}),
\end{equation*}
\noindent where $count(r, h)$ is the number of triples with $h$ as subject and $r$ as relation~\cite{DBLP:conf/dexa/MunozN17}.
\end{definition}

\begin{example}
Given a cardinality bound $\varphi_{\rel{hasParent}} = (0, 2)$, encoding the constraint \textit{``a person should have at most two parents''}, we would like to ensure that the embeddings learned by a neural link predictor yield predictions for the \rel{hasParent} relation within the boundaries.
In other words, we want to have the sum of probabilities over all possible parent entities of \ent{Edgar Allan Poe} precisely between zero and two.\footnote{Note that by considering a lower bound equals to zero, we can account for the possible incompleteness of the KG.}
We express this constraint over the triple $\tau = \triple{edgar\_allan\_poe}{hasParent}{t}$ as:
\begin{equation}\label{eq:card-bound}
 0 \leq \sum_{t \in \set{E}} p( y_{hrt} = 1 \mid \params) \leq 2,
\end{equation}
\noindent where the conditional probabilities $\forall t\in \set{E}$ are given by the neural link prediction model.
\end{example}
This term in \cref{eq:card-bound} expresses a supervision signal, not based on labelled data, that can be input to the training of neural link prediction models.
It is worth to mention that such cardinality boundaries can be provided by experts, gathered from literature~\cite{DBLP:conf/acl/MirzaRDW17}, or extracted from knowledge bases~\cite{DBLP:conf/dexa/MunozN17,DBLP:conf/wsdm/GalarragaRAS17}.
%

\section{Regularisation Based on Cardinality}\label{sec:regularizer}
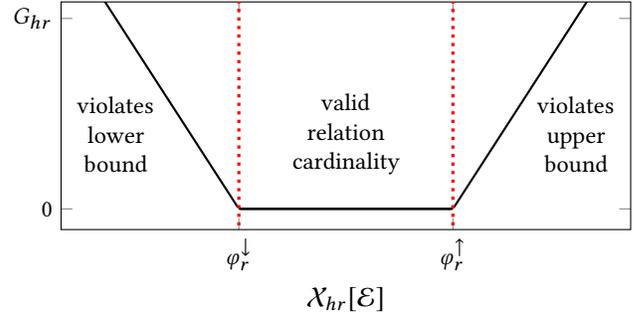
\begin{figure}[t]
 \centering
 \resizebox{\columnwidth}{!}{
    \begin{tikzpicture}[scale=\columnwidth/8.5cm]
    \begin{axis}[
        height=3cm,
        width=7.5cm,
        transform shape,
        scale only axis,
        ymax=25,
        label style={font=\large},
        xlabel={$\mathcal{X}_{hr}[\mathcal{E}]$},  
        xtick={30,70},
        xticklabels={$\varphi^{\downarrow}_{r}$,$\varphi^{\uparrow}_{r}$},
        ytick={0,23},
        yticklabels={0,$G_{hr}$},
        ]
        \addplot[domain=0:70, thick, samples=100] {max(0,30-x)};
        \addplot[domain=30:100, thick, samples=100] {max(0,x-70)};
        \node[align=center, text width=3cm] at (500,90) {valid \\ relation \\ cardinality};
        \node[align=center, text width=3cm] at (70,90) {violates \\ lower \\ bound};
        \node[align=center, text width=3cm] at (930,90) {violates \\ upper \\ bound};
        \draw[very thick, dotted, color=red] (300,-20) -- (300,400);
        \draw[very thick, dotted, color=red] (700,-20) -- (700,400);
    \end{axis}
    \end{tikzpicture}
 }
 \caption{Regularisation term $G_{hr}$ based on the bounds of a cardinality constraint $\varphi_{r} = (\bmin_{r}, \bmax_{r})$.}
 \label{fig:regularisation}
\end{figure}

%
In this section, we propose an approach to incorporate cardinality bounds in the training of neural link prediction models.
Specifically, we propose to leverage the available cardinality bounds, expressed as in \cref{eq:card-bound}, to define a regularisation term that encourages models to respect the available cardinality constraints.
Let $\Phi = \{\varphi_{r}=(\bmin_{r}, \bmax_{r}) \}_{r \in \set{R}}$ be the set of cardinality constraints for each relation in a given knowledge graph \kg, where $\bmin_{r}$ and $\bmax_{r}$ are the lower and upper bound for relation \rel{r}, respectively.
Given $r \in \set{R}$ and $h \in \set{E}$, let $\set{A}_{hr}[\set{E}] \triangleq \{\triple{h}{r}{t} : \forall t \in \set{E}\}$ be the set of all possible triples with relation $r$ and subject $h$, where the object $t$ was selected from $\set{E}$.
Following our toy example, assume that $r$ denotes the relation \rel{hasParent}, and $h$ denotes the entity \ent{edgar\_allan\_poe}.
Hence, we can take the set of possible triples to define the following hard constraint on the conditional probability of the triples in $\set{A}_{hr}[\set{E}]$:
\begin{equation}\label{eq:constraint}
\bmin_{r} \leq \left( \set{X}_{hr}[\set{E}] \triangleq \smashoperator{\sum_{x_{hrt} \in \set{A}_{hr}[\set{E}]}} p_{\params}(y_{hrt} = 1 \mid \params) \right) \leq \bmax_{r}.
\end{equation}
However, the inequality constraint in~\cref{eq:constraint} is impractical to incorporate directly in neural link predictors.
In this work, we propose a continuous relaxation of the constraint in~\cref{eq:constraint} to a \emph{soft constraint}, by defining a continuous and differentiable loss function that penalises violations of such a constraint.
Specifically, we define a function $G_{hr}$ that is strictly positive if the cardinality constraint for a given entity $h$ and relation $r$ is violated, and zero otherwise.
Given a cardinality constraint $\varphi_{r}$, the function $G_{hr}[\set{E};\Phi]$ (or $G_{hr}$ for simplicity) is defined as follows:
\begin{equation} \label{eq:regulariser}
\begin{aligned}
G_{hr}[\set{E};\Phi] = & \max ( 0, \bmin_{r} - \set{X}_{hr}[\set{E}])\ + \\
                  & \max ( 0, \set{X}_{hr}[\set{E}] - \bmax_{r} ).
\end{aligned}
\end{equation}
\Cref{fig:regularisation} shows the values of $G_{hr}$ (\cref{eq:regulariser}) based on $\set{X}_{hr}[\set{E}]$ and a cardinality bound $\varphi_{r}\in \Phi$.
Notice that for the general case where the upper bound corresponds to $\infty$ and lower bound to $0$, the loss $G_{hr}[\set{E};\Phi]$ vanishes.

Therefore, we define a cardinality-regularised objective function, denoted by $\set{L}^{C}(\params)$, for neural link prediction models: 
\begin{equation}\label{eq:regularised-loss}
\set{L}^{C}(\params) = \set{L}(\params) + \lambda \sum_{\Phi} G_{hr}[\set{E};\Phi],
\end{equation}
\noindent where $\lambda \in \Real_{+}$ weights the relative contribution of the regularisation term, and $\set{L}(\params)$ can be either the pairwise ranking loss or the logistic loss.
The regularised loss \cref{eq:regularised-loss} can be minimised using stochastic gradient descent (SGD)~\cite{robbins1951} in mini-batch mode, outlined in \cref{alg:sgd}.
Although our approach considers both upper and lower bounds, the latter cannot be meaningfully imposed in all cases.
For instance, given a constraint $\varphi_{spouse}=(1, 1)$, the regularisation term $G_{hr}[\set{E};\Phi]$ can yield inconsistent results if the knowledge graph is incomplete, and does not contain the \rel{spouse} link of every person.
In such cases, a zero lower bound can be used to address the knowledge graph incompleteness.
Our approach is intuitive and easy to implement for any neural link prediction model.
However, it is limited by the cost of computing the sum in~\cref{eq:constraint}: the set $\set{A}_{hr}[\set{E}]$ can easily grow in some KGs and become too expensive to obtain the sum of probabilities.
In the following section, we propose to use sampling techniques to overcome this problem by approximating the sum of probabilities.

\begin{algorithm}[t]
\caption{Learning the model parameters $\params$ via projected SGD} \label{alg:sgd}
\begin{algorithmic}[1]
	\Require{Observed facts $\set{D}^{+}$, epochs $\tau$, initial learning rate $\eta\in \Real$}
	\Ensure{Optimal model parameters $\params$ (see \cite{DBLP:journals/pieee/Nickel0TG16})}
	\State{Initialise embeddings $\vv{e}$ and $\vv{r}$ according to \cite{DBLP:journals/jmlr/GlorotB10}}
	\For{$i = 1,\ldots,\tau$}
	    \State{}\Comment{Build batch for training}
	    \State{$\set{T} \leftarrow$  sample a batch from $\set{D}^{+}$}
	    \State{$\set{B}^{+}\leftarrow \emptyset, \set{B}^{-}\leftarrow \emptyset$}
	    \For{$\tau^{+}=(\ent{h},\rel{r},\ent{t})\in \set{T}$}
	        \State{$\tau^{-} \in \set{C}(\ent{h},\rel{r},\ent{t})$} \Comment{Sample negative example}
	        \State{$\set{B}^{+}\leftarrow \set{B}^{+}\cup \{\tau^{+}\}, \set{B}^{-}\leftarrow \set{B}^{-}\cup \{\tau^{-}\}$}
	    \EndFor
		\State{}\Comment{Compute the gradient of the loss function $\set{L}$}
		\State{$g_{i} \leftarrow  \nabla \set{L}(\params)$ using $\set{B}^{+}$ and $\set{B}^{-}$}   
		\State{}\Comment{Model parameters update via gradient descent}
		\State{$\params_{i} \leftarrow \params_{i-1} - \eta_{i} g_{i}$} 
		\State{}\Comment{Projection step normalising all entity embeddings}
		\State{$\vv{e} \leftarrow \vv{e} / || \vv{e} ||, \; \forall e \in \set{E}$}
	\EndFor
	\State \textbf{return} $\params$
\end{algorithmic}
\end{algorithm}

\subsection{Lower Bound Estimation}\label{subsec:lower-bound}
We can sample a subset of all entities $\set{S} \subseteq \set{E}$ and obtain the following lower bound:
\begin{equation} \label{eq:lowerbound}
 \set{X}_{hr}[\set{S}] \leq \set{X}_{hr}[\set{E}].
\end{equation}
The tightness of the bound in~\cref{eq:lowerbound} is determined by the selection of the entities in $\set{S}$.
In this work, we consider \emph{uniform sampling}.
More specifically, a random set of indices $\set{S} \triangleq \{ i_1, \ldots, i_S \}$ is taken uniformly, where $i_s \in \{1, \ldots, C\}$, and form the following lower bound:
\begin{equation*}
 \smashoperator{\sum_{x_{hrt} \in \set{A}_{hr}[\set{S}]}} p(y_{hrt} = 1 \mid \params)  \leq \set{X}_{hr}[\set{E}],
\end{equation*}
where the sum is over all elements in \set{S} with no repetitions.
%

\subsection{Sum Estimation} \label{ssec:sum-estimation}
Instead of defining a lower bound to $\set{X}_{hr}[\set{E}]$, we can also approximate $\set{X}_{hr}[\set{E}]$ directly by \emph{sampling}. 
Let us consider a sum over a large collection of elements $Z \triangleq \sum_{c} z_c$.
We consider two standard methods for approximating sums via Monte Carlo estimates, namely Importance Sampling (IS) and Bernoulli Sampling~\cite{DBLP:conf/aistats/BotevZB17}.

\topic{Importance Sampling}
Based on the identity $Z = \sum_{c}\frac{q(c)z_c}{q(c)}$, a set of indices $\set{S}\equiv \{i_1,\ldots,i_S\}$ is selected from a distribution $q$, where $i_s\in \{1,\ldots,C\}$, and yielding the following approximation:
\begin{equation*}
 Z \approx \cfrac{1}{S} \sum_{s\in \mathcal{S}} \cfrac{z_{s}}{q(s)},
\end{equation*}
\noindent where $q(s)$ defines the probability of sampling $s$ from $\set{S}$.
%

\topic{Bernoulli Sampling}
An alternative to IS is Bernoulli Sampling (BS), considering the following identity:
\begin{equation*}
 Z = \sum_{c} z_c = \mathbb{E}_{\mathbf{s} \sim \mathbf{b}} \left( \sum_{c} \cfrac{s_c}{b_c} z_c \right),
\end{equation*}
where each independent Bernoulli variable $s_c\in \{0, 1\}$ denotes whether $z_{c}$ will be sampled or not, and $p(s_c = 1) = b_c$ is the probability of sampling $z_c$.
This leads to the following approximation:
\begin{equation*}
 Z \approx \sum_{c:s_c = 1} \cfrac{z_c}{b_c},
\end{equation*}
\noindent where the sum is computed over the components with non-zero elements in the vector $\mathbf{s}$.
Note that, when calculating an approximation to $Z$, IS relies on sampling with replacement, while BS relies on sampling without replacement.
By using our regularisation term with sampling, we add a time complexity $O(c d)$, where $c$ is the total number of (sampled) triples when computing the regularisation term, and $d$ the number of triples per batch.
Since $c$ can be smaller than the number of triples in a batch, we ensure that the time complexity of neural link predictors is not sensibly affected during training, and not affected at all at test time.
The proposed method does not increase the space complexity of the models, since the proposed regulariser does not change the number of model parameters.
%

\section{Evaluation}\label{sec:evaluation}
%
In this section, we investigate the benefits of cardinality regularisation for the state-of-the-art neural link prediction models.
We compare the performance of original and regularised losses in the link prediction task across different benchmark datasets, which are partitioned into train, validation and test set of triples (\cf~\cref{tab:datasets}).
%

\subsection{Evaluation Protocol}
The link prediction task consists of predicting a missing entity $h$ or $t$ when given a pair $(r,t)$ or $(h,r)$, respectively.
During testing, for each test triple \triple{h}{r}{t}, we replace the subject or object entity with all entities in the knowledge graph as corruptions~\cite{DBLP:conf/nips/BordesUGWY13}.
The evaluation then ranks the entities in descending order \wrt the scores calculated by a scoring function and gets the rank of the correct entity \ent{h} or \ent{t}.
We report results based on the ranks assigned to correct entities measured using mean reciprocal rank (MRR) and Hits@$n$ with $n\in \{1, 3, 5, 10\}$.\footnote{For MRR and Hits@$n$, the higher the better.}
During the ranking process some positive test triples could be ranked after another true triples, which should not be considered a mistake.
Therefore, the above metrics have two settings: \emph{raw} and \emph{filtered}~\cite{DBLP:conf/nips/BordesUGWY13}.
In the filtered setting, metrics are computed after removing all true triples appearing in train, validation, or test sets from the ranking, whereas in the raw setting they are not removed.
%

\subsection{Datasets}
Three widely used datasets for evaluating link prediction models are WordNet~\cite{DBLP:journals/cacm/Miller95}, Freebase~\cite{DBLP:conf/aaai/BollackerCT07}, and YAGO~\cite{DBLP:conf/cidr/MahdisoltaniBS15}.
In this work, we use four benchmark datasets generated from them: FB13, WN18, WN18RR and YAGO3-10.
The FB13 dataset~\cite{DBLP:conf/aaai/BordesWCB11} is a subset of Freebase containing 13 relation types and entities of type \textit{deceased\_people}, where entities appear in at least 4 relations and relation types at least 5,000 times.~\footnote{We use the corrected version by~\cite{DBLP:conf/nips/SocherCMN13} that contains only positive samples.}
We also use two datasets derived from WordNet, namely, WN18 and WN18RR.
These datasets contain hyponym, hypernym, and other lexical relations of English concepts and words.
It is known that WN18 contains \ca~72\% of redundant and inverse relations, which were removed in the WN18RR dataset~\cite{DBLP:conf/aaai/DettmersMS018}.
YAGO3-10 consists of entities in YAGO3 (mostly of the people type) linked with at least 10 relations, such as citizenship, gender and profession.
FB13, WN18RR, and YAGO3-10 datasets were shown to have no redundant or trivial triples~\cite{DBLP:conf/aaai/DettmersMS018}.
In~\cref{tab:datasets} we summarise the characteristics of each of the datasets.
\begin{table}[t]
\begin{center}
\resizebox{\columnwidth}{!}{%
\begin{tabular}{@{ } lccccc @{ }}
\toprule%
Dataset & $N_{r}$ & $N_{e}$ & $|\text{train}|$ & $|\text{validation}|$ & $|\text{test}|$ \\
\midrule%
FB13         & 13       & 81,065   & 350,517   & 5,000    & 5,000   \\
WN18         & 18       & 40,943   & 14,1442   & 5,000    & 5,000   \\
WN18RR       & 11       & 40,943   & 86,835    & 3,034    & 3,134   \\
YAGO3-10     & 37       & 123,182  & 1,079,040 & 5,000    & 5,000   \\
\bottomrule%
\end{tabular}
}%
\end{center}
\caption{Statistics for each of the datasets.}
\label{tab:datasets}
\end{table}
We mine the relation cardinality constraints from the training set of each dataset, following the algorithm proposed by~\citet{DBLP:conf/dexa/MunozN17} using the normalisation option but without filtering outliers.
\Cref{tab:cardinality-constraints-examples} gives examples of the cardinality constraints mined from each dataset.

\begin{table}[t]
    \centering
    \resizebox{.7\columnwidth}{!}{%
    \begin{tabular}{c|c}
        \toprule
        \rel{/people/person/place\_of\_birth}    & (0, 2)   \\
        \rel{/people/person/parents}           & (0, 2)   \\
        \rel{/people/person/gender}            & (1, 1)   \\
        \midrule
        \rel{\_hyponym}                         & (0, 380)  \\
        \rel{\_has\_part}                        & (0, 73)   \\
        \rel{\_hypernym}                        & (0, 4)    \\
        \midrule
        \rel{livesIn}                          & (0, 12)   \\
        \rel{hasGender}                        & (0, 1)    \\
        \rel{hasChild}                         & (0, 19)   \\
        \bottomrule
    \end{tabular}
    }
    \caption{Cardinality constraints extracted from FB13, WN18 (WN18RR) and YAGO3-10.}
    \label{tab:cardinality-constraints-examples}
\end{table}

\subsection{Results}
For our experiments, we re-implemented three models using the TensorFlow framework~\cite{DBLP:conf/osdi/AbadiBCCDDDGIIK16}, namely, ER-MLP~\cite{DBLP:conf/kdd/0001GHHLMSSZ14}, DistMult~\cite{DBLP:journals/corr/YangYHGD14a} and ComplEx~\cite{DBLP:conf/icml/TrouillonWRGB16} (which was recently proven to be equivalent to HolE~\cite{DBLP:conf/acl/HayashiS17}).
We compare the performance over the four benchmark datasets of each model as originally stated by their authors and with the cardinality regularisation term (\cf~\cref{eq:regularised-loss}).
As recommended by~\cite{DBLP:conf/icml/TrouillonWRGB16}, we minimise the logistic loss to train each model by using SGD, and AdaGrad~\cite{DBLP:journals/jmlr/DuchiHS11} to adaptively select the learning rate, initialised as $\eta_{0}=0.1$.
For each model and dataset, we selected hyperparameters maximising filtered Hits@10 on the validation set using an exhaustive grid search.
The evaluation of our approach is three-fold:
\begin{inparaenum}[(i)]
    \item we measure the effects of the regulariser in the link prediction task;
    \item we measure the effects of the different sampling techniques; and
    \item we measure the violations to the cardinality constraints before and after regularisation.
\end{inparaenum}
To reduce the search space, during the grid search in (i) we fix the sampling technique to uniform.
In (ii), we use the best model identified in (i) to study the effect of different sampling techniques, whilst in (iii) we use the overall best model per dataset.
%

\topic{Link Prediction}
We train each model for 1,000 epochs with a mini-batches approach over the training set of each dataset, generating two negative examples per positive triple in each batch.
We set $\lambda=0$ to obtain the performance results of original models (without regularisation), and use uniform sampling with sizes $\mu\in \{10, 100\}$, $\omega\in \{10, 100, 1000\}$ of subjects and objects.\footnote{We identified via independent experiments that larger values for $\mu$ do not yield performance improvements.} 
\Cref{tab:link-prediction-fb13-wn18-wn18rr,tab:link-prediction-yago} show the link prediction results, confirming that in general our cardinality-based regularisation term helps to improve (or at least maintain) the performance of the original ER-MLP, DistMult and ComplEx models across all datasets.
The only exception we observed is ComplEx over YAGO3-10, where the model without the regularisation term reaches better Hits@10 and MRR.
We believe that a reason for this is that constraining a lower bound on the sum of probabilities may not be the best technique to use when the number of entities is very large.
In our experiments we also compare two alternative approaches, namely estimating the sum of probabilities via IS and BS.
ER-MLP and DistMult models benefit the most across all datasets with improvements of up to 36\% in MRR.
ComplEx shows to be the overall best performing model outperforming ER-MLP (up to 20x in WN18RR) and DistMult in every dataset and evaluation metric.
Still, ComplEx benefits from the regularisation term in most of the datasets.
Although we did not perform a thorough search of the hyperparameters space to reach state-of-the-art performance, the results prove the advantages of our approach.
\begin{table*}[ht]
\renewcommand\extrarowheight{1.3pt}
\begin{center}
\resizebox{\textwidth}{!}{%
\begin{tabular}{l
C{1cm}C{1cm}C{1cm}C{1cm}C{1cm}C{.1cm}  
C{1cm}C{1cm}C{1cm}C{1cm}C{1cm}C{.1cm}  
C{1cm}C{1cm}C{1cm}C{1cm}C{1cm}C{.1cm}  
}
\toprule
& \multicolumn{5}{c}{\bf FB13}   &   
& \multicolumn{5}{c}{\bf WN18}   &   
& \multicolumn{5}{c}{\bf WN18RR}
\\
\cline{2-6}\cline{8-12}\cline{14-18}
& \multicolumn{4}{c}{Hits@$n$} & \multirow{2}{*}{MRR} &
& \multicolumn{4}{c}{Hits@$n$} & \multirow{2}{*}{MRR} &
& \multicolumn{4}{c}{Hits@$n$} & \multirow{2}{*}{MRR}
\\
\cline{2-5}\cline{8-11}\cline{14-17}
{\bf Method} 
& 1 & 3 & 5 & 10 & & \ 
& 1 & 3 & 5 & 10 & \ & 
& 1 & 3 & 5 & 10 & \
\\
\midrule
ER-MLP & 
4.40        & 7.55        & 9.14        & 11.82       & 6.94        &     & 
21.64       & 37.30       & 44.94       & 56.52       & 33.02       &     & 
1.84        & 3.29        & 4.10        & 5.31        & 3.10        
\\
ER-MLP$^{C}$ & 
\bb{5.13}        & \bb{8.36}          & \bb{10.29}        & \bb{12.75}        & \bb{7.78}         &     &
\bb{32.01}       & \bb{51.54}         & \bb{60.54}        & \bb{70.85}        & \bb{45.01}        &     &
\bb{2.22}        & \bb{4.29}          & \bb{5.42}         & \bb{7.31}         & \bb{3.98} 
\\
\midrule
DistMult &
18.07        & 29.29        & 32.94        & 37.01        & 24.92       &     &
64.46        & 87.47        & 90.66        & 93.49        & 76.62       &     &
38.93        & 43.49        & 45.93        & 49.63        & 42.46       
\\
DistMult$^{C}$ &
\bb{18.10}        & \bb{29.45}        & \bb{33.07}        & \bb{37.02}         & \bb{25.00}        &     & 
\bb{65.01}        & \bb{87.53}        & \bb{90.71}        & \bb{93.44}         & \bb{76.93}        &     & 
\bb{39.10}        & \bb{44.13}        & \bb{46.30}        & \bb{49.81}         & \bb{42.84}
\\
\midrule
ComplEx &
\best{\bb{25.08}} & 31.64        & 34.00        & 36.90        & \best{\bb{29.41}} &     &
88.33        & 93.05        & 94.14        & 95.07        & 90.96        &     & 
40.87        & \best{\bb{46.25}} & \best{\bb{48.55}} & \best{\bb{51.15}} & 44.52      
\\
ComplEx$^{C}$ &
24.89        & \best{\bb{31.78}} & \best{\bb{34.10}} & \best{\bb{37.16}}  & 29.36        &     &
\best{\bb{88.66}} & \best{\bb{93.27}} & \best{\bb{94.21}} & \best{\bb{95.21}}  & \best{\bb{91.20}} &     & 
\best{\bb{41.10}} & 46.06        & 48.13        & 51.09         & \best{\bb{44.57}} 
\\
\bottomrule
\end{tabular}
}%
\end{center}
\caption{Link prediction results (Hits@$n$ and Mean Reciprocal Rank, filtered setting) on FB13, WN18 and WN18RR. In bold the best results comparing both original and cardinality loss, and highlighted is the best value per evaluation metric across all models.}
\label{tab:link-prediction-fb13-wn18-wn18rr}
\end{table*}
\begin{table}[ht]
\begin{center}
\resizebox{.85\columnwidth}{!}{%
\begin{tabular}{l
C{1cm}C{1cm}C{1cm}C{1cm}C{1cm}
}
\toprule
& \multicolumn{5}{c}{\bf YAGO3-10} \\
\cline{2-6}
& \multicolumn{4}{c}{Hits@$n$} & \multirow{2}{*}{MRR} \\
\cline{2-5}
{\bf Method} 
& 1 & 3 & 5 & 10 & \ \\
\midrule
ER-MLP & 
2.22        & 6.09         & 9.59         & 16.01        & 6.83        \\ 
ER-MLP$^{C}$ & 
\bb{2.33}        & \bb{6.16}         & \bb{9.65}         & \bb{16.54}        & \bb{6.95}         \\
\midrule
DistMult &
6.75        & 14.33        & 18.86        & 26.51        & 13.33        \\
DistMult$^{C}$ &
\bb{7.03}        & \bb{14.53}        & \bb{19.12}        & \bb{26.66}        & \bb{13.59}        \\
\midrule
ComplEx &
7.12        & \best{\bb{15.61}} & \best{\bb{20.76}} & \best{\bb{29.11}} & 14.33        \\
ComplEx$^{C}$ &
\best{\bb{7.56}} & 15.10        & 20.30        & 29.01        & \best{\bb{14.47}}  \\
\bottomrule
\end{tabular}
}%
\end{center}
\caption{Link prediction results (Hits@$n$ and Mean Reciprocal Rank, filtered setting) on YAGO3-10. In bold the best results comparing both original and cardinality loss, and highlighted is the best value per evaluation metric across all models}
\label{tab:link-prediction-yago}
\end{table}
%

\topic{Sampling techniques}
To approximate the sum of probabilities we test both Importance Sampling and Bernoulli Sampling, and consider hyperparameters $\mu\in \{10,$ $50,$ $100\}$ and $\omega\in \{10,$ $50,$ $100,$ $500,$ $1000\}$.
Starting from the best ComplEx models learned above, we tune the sampling technique for each of the datasets.
Results are shown in~\cref{tab:sampling-techniques-results}.
In general, all sampling techniques work well and there is no \textit{one-size-fits-all} solution: it depends on the dataset.
(Information about properties of the data that benefit one of the samplings can be used, and custom sampling is also supported.)
YAGO3-10 shows the biggest improvement of 6\% in MRR using BS compared with the results in~\cref{tab:link-prediction-yago}.
This improvement might be correlated to the advantage of BS to handle the large number of entities in YAGO3-10.
For FB13, WN18, and WN18RR we see smaller improvements in MRR and Hits@10 compared to the results in~\cref{tab:link-prediction-fb13-wn18-wn18rr}.
Differences in results for uniform sampling compared to the results in~\cref{tab:link-prediction-fb13-wn18-wn18rr} are also attributed to the expanded hyperparameters space with more sampling sizes than previously.
\begin{table}[t]
\begin{center}
\resizebox{\columnwidth}{!}{%
\begin{tabular}{cl
C{1cm}C{1cm}C{1cm}C{1cm}C{1cm}
}
\toprule
& & \multicolumn{4}{c}{Hits@$n$} & \multirow{2}{*}{MRR} \\
\cline{3-6}
{\bf Dataset} & {\bf Sampling} & 1 & 3 & 5 & 10 & \ \\
\midrule
\multirow{3}{*}{FB13} & Uniform & 
25.84        & 31.85        & \best{34.19} & \best{37.26} & 29.89        \\
& Importance & 
25.17        & 31.36        & 34.36        & 36.18        & 29.18        \\
& Bernoulli & 
\best{25.92} & \best{31.86} & 34.11        & 37.18        & \best{29.97} \\
\hline
\multirow{3}{*}{WN18} & Uniform &
88.98        & \best{93.66} & \best{94.84} & 95.98        & \best{92.12} \\
& Importance &
88.97        & 93.64        & 94.73        & \best{96.08} & 91.10        \\
& Bernoulli &
\best{89.05} & 93.57        & 94.67        & 95.94        & 91.09       \\
\hline
\multirow{3}{*}{WN18RR} & Uniform &
41.27        & 46.57        & 48.58        & \best{51.51} & 44.87       \\
& Importance &
41.09        & 46.68        & \best{48.81} & 51.50        & 44.78       \\
& Bernoulli &
\best{41.54} & \best{46.79} & 48.68        & 51.42        & \best{45.04} \\
\hline
\multirow{3}{*}{YAGO3-10} & Uniform &
8.32        & 15.52        & \best{20.92} & 29.29        & 15.30        \\
& Importance &
8.23        & 15.71        & 20.70        & 29.49        & 15.28        \\
& Bernoulli &
\best{8.48} & \best{15.74} & 20.82        & \best{29.50} & \best{15.42} \\
\bottomrule
\end{tabular}
}%
\end{center}
\caption{Link prediction results (Hits@$n$ and Mean Reciprocal Rank, filtered setting) for the best ComplEx model using different sampling techniques.}
\label{tab:sampling-techniques-results}
\end{table}
%

\begin{figure}[t]
    \centering
    \subfigure[\rel{imports} (0, 6)]{%
        \label{fig:violin-plot-imports}%
        \includegraphics[width=0.5\columnwidth]{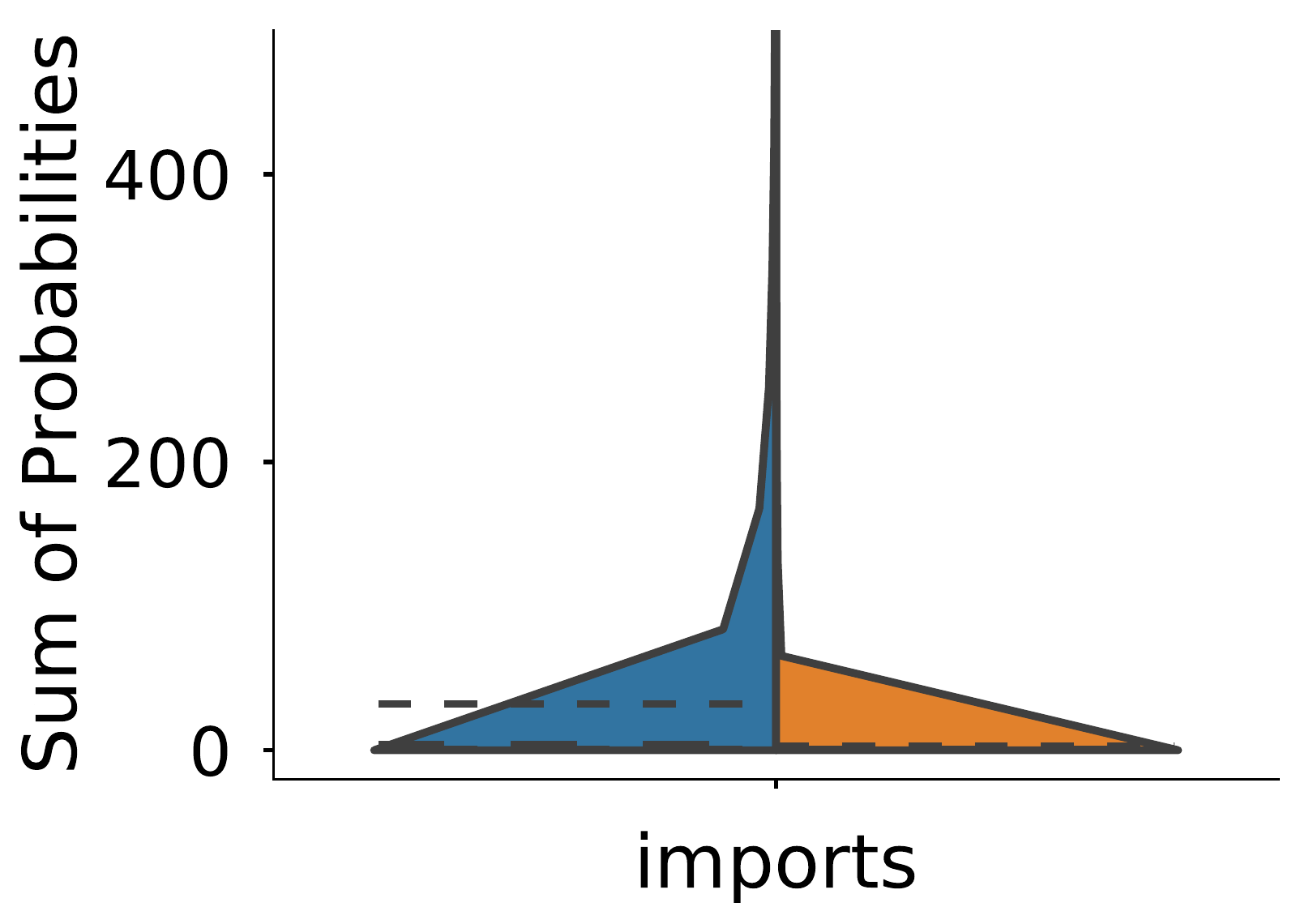}}%
    \subfigure[\rel{hasWebsite} (0, 2)]{%
        \label{fig:violin-plot-haswebsite}%
        \includegraphics[width=0.5\columnwidth]{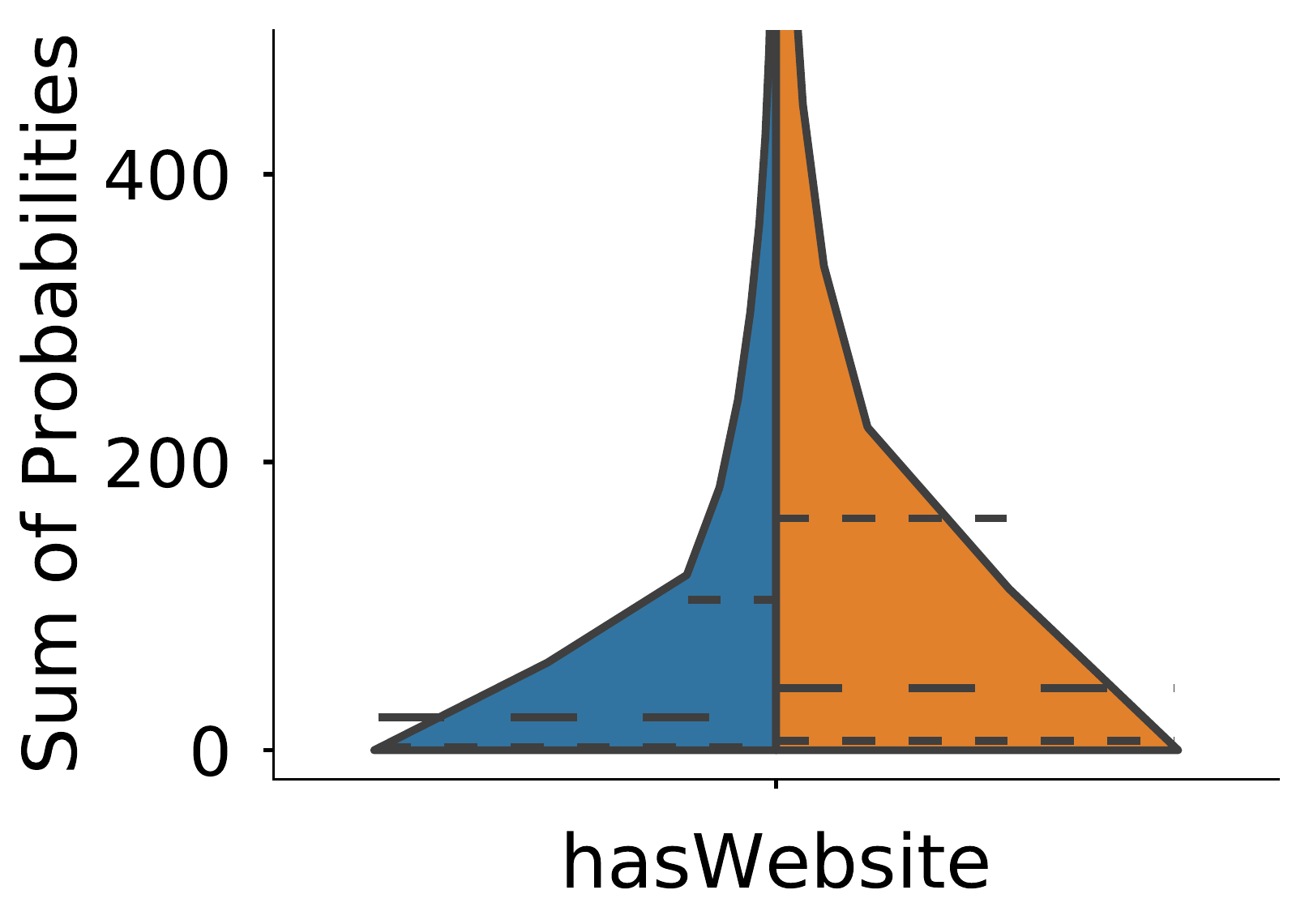}}%
    \\
    \subfigure[\rel{hasAcademicAdvisor} (0, 4)]{%
        \label{fig:violin-plot-hasacademicadvisor}%
        \includegraphics[width=0.5\columnwidth]{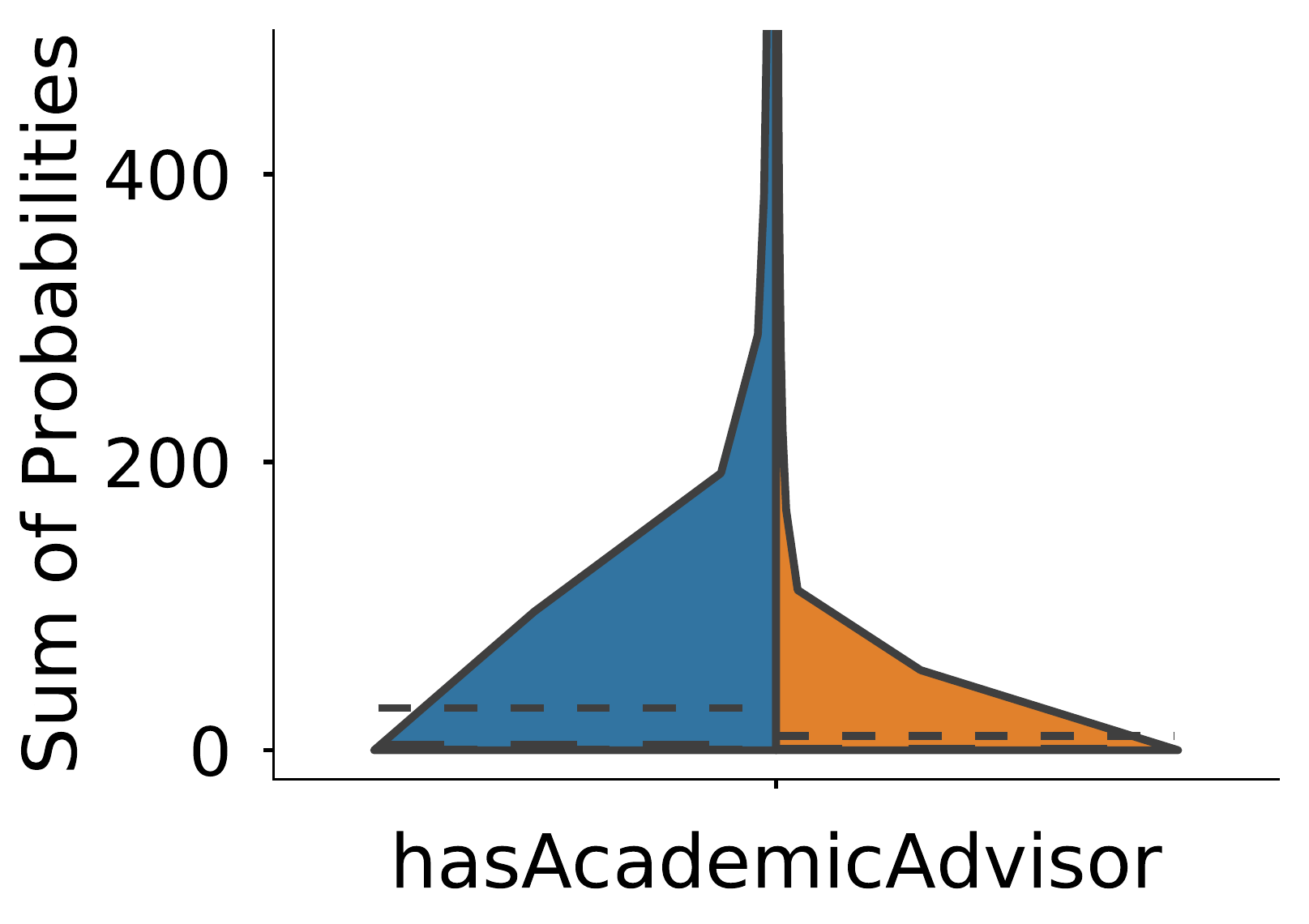}}%
    \subfigure[\rel{hasChild} (0, 19)]{%
        \label{fig:violin-plot-haschild}%
        \includegraphics[width=0.5\columnwidth]{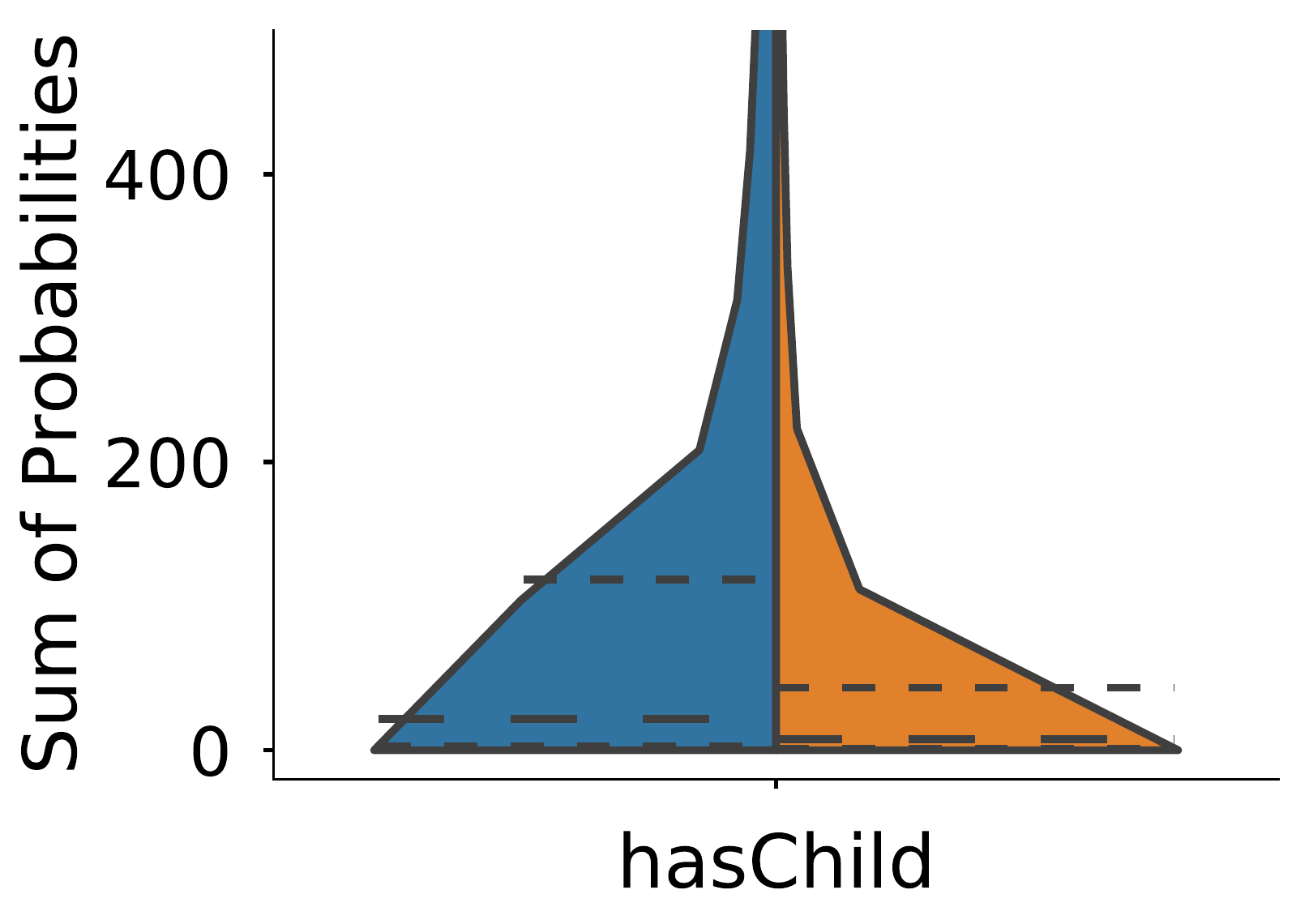}}%
    \caption{Changes in the distribution of $\set{X}_{hr}[\set{E}]$ without (left, in blue) and with (right, in orange) regularisation using ER-MLP in YAGO3-10. Horizontal lines correspond to quartiles.}
    \label{fig:violin-plots}
\end{figure}

\begin{figure}[t]
    \centering
    \resizebox{0.9\columnwidth}{!}{
    
    \definecolor{color1}{RGB}{5,113,176}
    \definecolor{color2}{RGB}{244,165,130}
    \definecolor{color3}{RGB}{146,197,222}
    \definecolor{color4}{RGB}{202,0,32}

    \begin{tikzpicture}[font=\sffamily,scale=\columnwidth/9.5cm]
    \begin{axis}[
        colormap/cool,
        xlabel={Regularisation parameter $\lambda$},
        ylabel={Avg. mean of $\mathcal{C}_{hr}[\mathcal{E}]$},
        xlabel near ticks,
        ylabel near ticks,
        grid=major,
        axis y line*=left,
        every axis plot/.append style={ultra thick},
        xmin=-0.3,
        xmax=5.3,
        ymin=0,
        xtick={0,1,2,3,4,5},
        xticklabels = {$0$,$1^{-4}$,$1^{-3}$,$1^{-2}$,$1^{-1}$,$1.0$},
    ]
    \addplot[solid,mark=triangle*,mark options={solid,scale=2},color=color3] table [id=exp]{data/wn18_violations.data};
    \legend{};
    \end{axis}
    
    \begin{axis}[
        colormap/cool,
        ylabel=Hits@10 score,
        ylabel near ticks,
        hide x axis,
        axis y line*=right,
        every axis plot/.append style={ultra thick},
        xmin=-0.3, 
        xmax=5.3,
        ymin=0,
        ymax=1.1,
    ]
    \addplot[dashed,mark=triangle*,mark options={solid,scale=2},color=color4] table [id=exp]{data/wn18_mrr.data};
    \end{axis}
    \end{tikzpicture}
    
    }
    \caption{Influence of the regularisation weight over the average mean of $\mathcal{X}_{hr}[\mathcal{E}]$ (solid blue line) and Hits@10 (dashed red line) in WN18 with ComplEx.}
    \label{fig:lambda-test}
\end{figure}
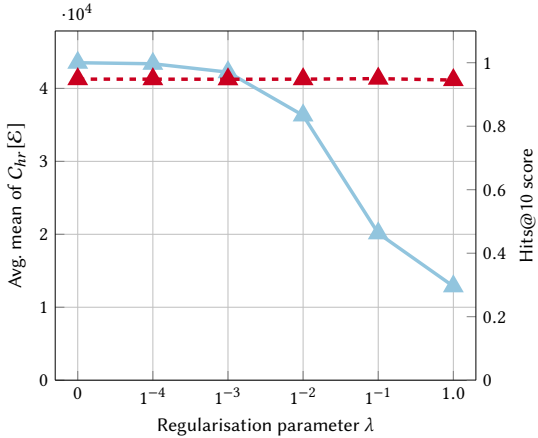

\topic{Cardinality Violations in KGs}
We have shown that our regulariser is beneficial for the link prediction task, but, more importantly, the predictions that violate the cardinality constraints are significantly reduced.
\Cref{fig:violin-plots} shows the changes on the distribution of $\set{X}_{hr}[\set{E}]$ in four relation cases for ER-MLP in YAGO3-10---one of the most benefited settings.
\Cref{fig:violin-plot-imports,fig:violin-plot-hasacademicadvisor,fig:violin-plot-haschild} illustrate positive impacts of the regularisation.
We observed that the regulariser decreases the median and long-tail distribution above the third quartile for (almost) every relation, making predictions more accurate.
For example, in relation \rel{imports} ($\varphi=(0, 6)$) the mean of $\set{X}_{hr}[\set{E}]$ is reduced by 78\%, meaning less violations.
Conversely, the biggest negative impact was in relation \rel{hasWebsite} ($\varphi=(0, 2)$, \cref{fig:violin-plot-haswebsite}), where violations were increased by 65\%.
Both constraint are equally restrictive over the number of objects but they differ on their range.
For the former, the objects are entities with links to other entities, while in the latter objects are literals (URLs) with no further links.
The prediction of literals is a known problem for neural link predictors as there are not many links to other entities~\cite{DBLP:journals/corr/abs-1709-04676}.
Following the DistMult example using the constraint $\varphi_{\rel{hasParent}} = (0, 2)$, \cref{tab:toy-example-regularized} shows the predictions for parents of \ent{Edgar Allan Poe}.
There are less predictions with high probability and a correct, but previously missing, entity \ent{David Poe Jr.} is now scored with a high probability proving the effectiveness of regularisation.
\begin{table}[t]
\centering
\resizebox{0.85\columnwidth}{!}{%
\begin{tabular}{lc}
\toprule
\multicolumn{1}{c}{\bf Triple}            & {\bf Probability} \\
\midrule
\triple{edgar}{hasParent}{eliza\_poe}     & 0.861  \\
\triple{edgar}{hasParent}{maria\_poe}     & 0.854  \\
\triple{edgar}{hasParent}{david\_poe\_jr} & 0.815  \\
\bottomrule
\end{tabular}
}
\caption{Predictions with probability $>0.8$ for \triple{edgar\_allan\_poe}{hasParent}{?} by DistMult when imposing the cardinality regulariser.}
\label{tab:toy-example-regularized}
\end{table}
We did not note any major difference in results between tight and loose cardinality bounds, or between constraints for relations with few and many instances.
Finally, \cref{fig:lambda-test} shows the effects of using different regularisation weights $\lambda\in \{0, 0.0001, 0.001, 0.01,\allowbreak 0.1, 1.0\}$ over the values of average mean of $\mathcal{X}_{hr}[\mathcal{E}]$ and Hits@10 across relations in WN18RR.
As $\lambda$ grows, Hits@10 suffers small changes and the average mean of $\mathcal{X}_{hr}[\mathcal{E}]$ decreases.
This shows that the regularisation term does not affect negatively Hits@10 (a common evaluation metric) and helps to decrease the number of violations to the cardinality constraints.
%

\section{Related Work}\label{sec:rel-work}
%
Early works in neural link prediction (\eg, TransE~\cite{DBLP:conf/nips/BordesUGWY13}, RESCAL~\cite{DBLP:conf/icml/NickelTK11}, DistMult~\cite{DBLP:journals/corr/YangYHGD14a}) learn the representations of all entities and relations in the knowledge base by fitting simple scoring functions on the triples in the knowledge graph.
Recently, research focused on either
\begin{inparaenum}[(i)]
    \item generating more elaborated scoring functions that better capture the nature of each of the relations, or
    \item improving existing models with background knowledge~\cite{DBLP:journals/tkde/WangMWG17}.
\end{inparaenum}
The former includes HolE~\cite{DBLP:conf/aaai/NickelRP16}, where the scoring function is inspired by cognitive models of associative memory; ComplEx~\cite{DBLP:conf/icml/TrouillonWRGB16} that uses complex-valued embeddings to model asymmetric relations; and ConvE~\cite{DBLP:conf/aaai/DettmersMS018} that builds a multi-layer convolutional network.
The latter is characterised by the incorporation of additional information such as entity types, relation paths, and logical rules.
We refer the readers to~\cite{DBLP:journals/pieee/Nickel0TG16,DBLP:journals/tkde/WangMWG17} for a deeper review of neural link predictors.
Our work aligns with the second category that focuses on adding background knowledge.
Almost every paper incorporating background knowledge agree that such prior knowledge improves link prediction models~\cite{DBLP:conf/acl/GuoWWWG15,DBLP:conf/sac/MinervinidFE16,DBLP:conf/uai/MinerviniDRR17,DBLP:conf/pkdd/MinerviniCMNV17,DBLP:journals/tkde/GuoWWWG17,DBLP:conf/acl/WangWGD18}. However, none of them has considered integrity constraints such as cardinality.
\citeauthor{DBLP:conf/dexa/MunozN17} mine cardinality constraints from knowledge graphs, and suggest their use to improve the accuracy of link prediction models.
In a similar vein, \citeauthor{DBLP:conf/wsdm/GalarragaRAS17} use fine-grain cardinality information to prune `unnecessary' predictions. However, this is done only after the predictions are generated.
In~\cite{DBLP:conf/wsdm/ZhangCZCY17}, a single cardinality bound (one-to-one, one-to-many or many-to-many) is imposed in link prediction over single-relational graphs (such as organisational charts), which differs from the multi-relational nature of knowledge graphs.
%

\section{Conclusions}\label{sec:conclusions}
%
In this paper, we presented a cardinality-based regularisation term for neural link prediction models.
The regulariser incorporates background knowledge in the form of relation cardinality constraints that hitherto have been ignored by neural link predictors.
The incorporation of this regularisation term in the loss function significantly reduces the number of violations produced by models at prediction time, enforcing the number of predicted triples with high probability for each relation to satisfy cardinality bounds.
Experimental results show that the regulariser consistently improves the quality of the knowledge graph embeddings, without affecting the efficiency or scalability of the learning algorithms.
\begin{acks}
This work was partially supported by the TOMOE project funded by Fujitsu Laboratories Ltd., Japan and Insight Centre for Data Analytics at National University of Ireland Galway (supported by the Science Foundation Ireland (SFI) under Grant Number SFI/12/RC/2289).
\end{acks}

\balance

\bibliographystyle{ACM-Reference-Format}
\bibliography{references}
\end{document}